\documentclass[10pt,twocolumn,letterpaper]{article}

\usepackage[pagenumbers]{iccv} 

\newcommand{\TODO}[1]{\textbf{\color{red}[TODO: #1]}}
\renewcommand{\TODO}[1]{}

\usepackage{amssymb}
\usepackage{amsfonts}

\usepackage{amsmath}
\usepackage{algorithm}
\usepackage{algpseudocode}
\usepackage{multirow}
\usepackage{pifont}
\newcommand{\cmark}{\ding{51}}%
\newcommand{\xmark}{\ding{55}}%
\usepackage{graphicx}
\usepackage{booktabs}
\usepackage{makecell}
\usepackage{tabularx}
\usepackage{balance}
\usepackage{ulem}
\usepackage{bm}

\usepackage{indentfirst}

\definecolor{iccvblue}{rgb}{0.21,0.49,0.74}

\usepackage[pagebackref,breaklinks,colorlinks,allcolors=iccvblue]{hyperref}

\definecolor{LightCyan}{rgb}{0.88,1,1}

\title{Towards Accurate and Efficient 3D Object Detection for Autonomous Driving: \\A Mixture of Experts Computing System on Edge

}
\vspace{-2mm}

\small{
\author{Linshen Liu$^{1,*}$~
Boyan Su$^{1,*}$~ 
Junyue Jiang$^{1}$ ~
Guanlin Wu$^{1}$ \\
Cong Guo$^{2}$ ~
Ceyu Xu$^{3}$ ~
Hao Frank Yang$^{1,\dagger}$
\\
{$^{1}$Johns Hopkins University \quad $^{2}$Duke University \quad $^{3}$HKUST}\\
{\tt\small \{lliu148, bsu11\}@jh.edu, haofrankyang@jhu.edu}
}}
\begin{document}

\maketitle
\renewcommand{\thefootnote}{\fnsymbol{footnote}}
\footnotetext{$^{*}$Equal contribution, $^{\dagger}$Corresponding author.}

\begin{abstract}

\vspace{-2mm}

This paper presents \textbf{E}dge-based \textbf{M}ixture of Experts (MoE) \textbf{C}ollaborative \textbf{C}omputing (EMC2), an optimal computing system designed for autonomous vehicles (AVs) that simultaneously achieves low-latency and high-accuracy 3D object detection. Unlike conventional approaches, EMC2 incorporates a scenario-aware MoE architecture specifically optimized for edge platforms. By effectively fusing LiDAR and camera data, the system leverages the complementary strengths of sparse 3D point clouds and dense 2D images to generate robust multimodal representations. To enable this, EMC2 employs an adaptive multimodal data bridge that performs multi-scale preprocessing on sensor inputs, followed by a scenario-aware routing mechanism that dynamically dispatches features to dedicated expert models based on object visibility and distance. In addition, EMC2 integrates joint hardware-software optimizations, including hardware resource utilization optimization and computational graph simplification, to ensure efficient and real-time inference on resource-constrained edge devices. Experiments on open-source benchmarks clearly show the EMC2 advancements as an end-to-end system. On the KITTI dataset, it achieves an average accuracy improvement of 3.58\% and a 159.06\% inference speedup compared to 15 baseline methods on Jetson platforms, with similar performance gains on the nuScenes dataset, highlighting its capability to advance reliable, real-time 3D object detection tasks for AVs. The official implementation is available at \url{https://github.com/LinshenLiu622/EMC2}.

\vspace{-4mm}
\end{abstract}
\section{Introduction}
\label{sec:intro}


Traffic safety is a fundamental concern for both human drivers and Autonomous Driving Systems (ADS). Within an ADS, the perception module plays a pivotal role by serving as the "eyes" of the autonomous vehicle (AV), sensing the surrounding environment and delivering critical information to downstream modules such as motion planning, control and safety guarantees. Existing conceptual safety of an AV perception system is generally defined by two core dimensions: accuracy and efficiency. Accuracy ensures reliable object detection and tracking capabilities, enabling a credible understanding of the environment. Efficiency, characterized by low perception delay, is equally essential, as real-time high-level planning and control rely on fast information processing.  
Perception delay, defined as the sum of data acquisition, computation, and algorithm processing time, directly affects an ADS's responsiveness to dynamic environments. Research indicates that an ADS must process sensory inputs and issue feedback control commands within $100$ milliseconds to guarantee prompt and safe vehicle operations~\citep{lin2018thearchetectural}. \textbf{Detection precision and perception delay together define a critical safety boundary: a precise but sluggish system cannot react in time, while a fast but inaccurate system may lead to unsafe decisions.}

\begin{figure}[ht]
    \centering
    \vspace{-4mm}
    \includegraphics[width=0.86\linewidth]{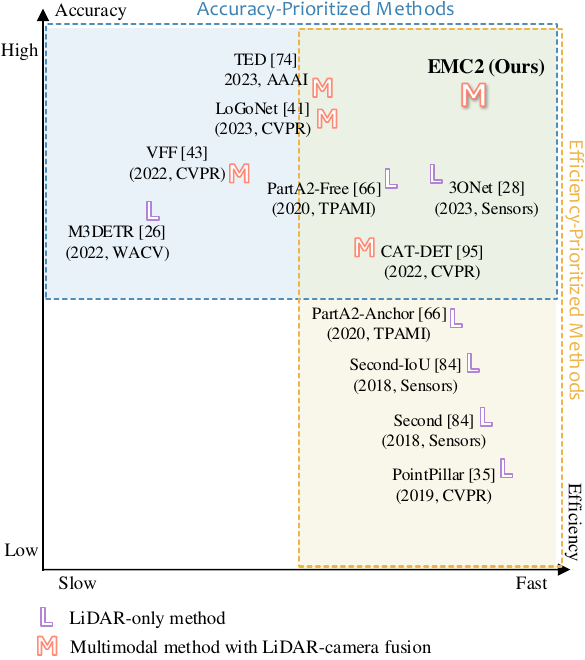}
    \caption{\textbf{Comparison of 3D object detection models in accuracy and inference efficiency on the KITTI and nuScenes dataset.} Most existing methods improve accuracy at the expense of inference efficiency. The proposed $\mathrm{EMC2}$ achieves win-win performance by customizing MoE for autonomous driving.}
    \vspace{-6mm}
    \label{fig:sota}
\end{figure}


As illustrated in Fig.~\ref{fig:sota}, achieving both high accuracy and low latency remains a fundamental challenge. Existing approaches often face inherent trade-offs between these objectives due to limited computational resources, insufficient adaptability of perception systems, and the inherent complexity of dynamic driving scenarios. Urban traffic environments require low-latency and accurate sensing capabilities to handle complex, densely interactive surroundings, while highway scenarios demand reliable detection of distant objects and prompt responses to unexpected events. Moreover, even for the same object, recognition priorities may vary dynamically within a scene depending on distance and scene complexity, posing notable challenges for single-modality models in balancing efficiency and accuracy~\citep{xiao2020multimodal}. Multimodal models, such as large-scale CNN and Transformer-based architectures that integrate LiDAR, radar, and camera data~\citep{deng2020voxelrcnn}, leverage the complementary advantages of diverse sensors to enhance perception accuracy. Nevertheless, these approaches~\citep{liao2024vlm2scene} often impose substantial hardware overhead and exhibit limited adaptability, particularly when deployed on resource-constrained edge platforms~\citep{zhang2015optimizing}. Even with the assistance of low-power machine learning compilation frameworks~\citep{chen2018tvm}, such systems may still execute redundant computations under straightforward scenarios, introducing additional delays that undermine timely safety-critical decision-making.

Nevertheless, multimodal approaches provide a natural foundation for addressing the accuracy challenges faced by ADS. Rather than relying on a single integrated model designed to operate uniformly across diverse driving scenarios, the MoE framework~\citep{cai2021ace} dynamically activates specialized sub-models, or experts, based on input characteristics and prior knowledge. MoE has demonstrated considerable effectiveness in large-scale applications, such as Large Language Models (LLMs) for natural language processing~\citep{li2025uni,li2024cumo}, where it routes different tokens to appropriate experts, reducing redundant computations without sacrificing efficiency~\cite{cai2024survey}. As model complexity grows, MoE enables models to scale their capacity while keeping inference hardware costs manageable. Deploying MoE in ADS, however, poses notable challenges~\citep{shi2024time}, including strict real-time constraints, the need for multimodal data processing, and seamless end-to-end autonomous feedback control. Successful deployment requires efficient data preprocessing, adaptive modality selection, and robust expert routing mechanisms.

By leveraging diverse sensor data and a scenario-adaptive MoE dispatcher, we dynamically assign expert sub-models based on scenario data and prior statistical knowledge, providing an efficient, energy-efficient, and safe solution for ADS. To this end, we introduce an \textbf{E}dge-based \textbf{M}oE \textbf{C}ollaborative \textbf{C}omputing ($\mathrm{EMC2}$) framework -- an algorithm–hardware–software co-design solution that not only improves accuracy through complex multimodal experts but also effectively reduces latency by utilizing an adaptive MoE dispatcher and lightweight single-modal experts. Our main contributions are summarized as follows:

\noindent \textbf{1) A Multimodal 3D Object Detection Solution with MoE.} The proposed $\mathrm{EMC2}$ system achieves both high detection accuracy and inference efficiency, outperforming $15$ existing one-size-fits-all baseline models. Designed for real-world ADS applications, $\mathrm{EMC2}$ is optimized for deployment on resource-constrained platforms, making it more suitable for edge computing environments than generic alternatives.

\vspace{1mm}
    
\noindent \textbf{{2) Robust and Efficient Multimodal Perception through Expert Collaboration and System-Level Optimization.}} The proposed $\mathrm{EMC2}$ framework incorporates scenario-specific experts and employs an adaptive expert selection strategy to meet heterogeneous perception demands. To mitigate the training instability commonly observed in MoE-based systems, we design a hierarchical optimization scheme that combines expert-level and global $\mathrm{EMC2}$ backpropagation for effective collaborative learning. Furthermore, a distribution-aware resampling technique enhances robustness under real-world data imbalance. In addition to these designs, a tailored multiscale pooling module improves the efficiency of LiDAR–camera fusion, ensuring both efficient and accurate multimodal perception.


\noindent \textbf{3) System-Level Optimization for Edge Inference.} The original PyTorch-based MoE inference framework~\citep{he2021fastmoe} suffers from inefficient management of memory and computational resources, leading to high latency on edge devices. To overcome these challenges, we propose an algorithm–hardware–software co-optimization strategy within $\mathrm{EMC2}$, which incorporates efficient memory management techniques to improve L1/L2 cache hit rates and applies computational graph fusion to eliminate redundant data loading across layers. These combined optimizations notably improve inference efficiency.

\vspace{1mm}


\noindent \textbf{4) State-of-the-art (SOTA) Accuracy and Efficiency Combined.} On the KITTI dataset, $\mathrm{EMC2}$ achieves 3.28\% higher accuracy for pedestrians and 6.03\% for bicycles under hard-level detection challenge compared to 15 baseline models. On the \textit{nuScenes} dataset, it yields a 3.9\% increase in mAP and a 1.8\% boost in NDS. In addition, $\mathrm{EMC2}$ achieves a 159.06\% reduction in inference latency on the Jetson platform for KITTI, supporting its suitability for real-time edge deployment.
\vspace{1mm}

\vspace{-3mm}

\section{Related Work}
\label{sec:related_work}
\vspace{-1mm}
\textbf{LiDAR-only Method.} The active 3D sensing of LiDAR makes it dominant in ADS 3D detection~\citep{qi2020imvotenet, shi2019parta2, alex2019pointpillars}. LiDAR-based methods process point clouds via voxelization~\citep{zhou2018voxelnet}, pillarization~\citep{alex2019pointpillars}, or direct raw input~\citep{shi2020pvrcnn}. Anchor-based heads, used for region proposal~\citep{zhou2018voxelnet}, and center-based representations\citep{yin2021center} are widely adopted. These methods achieve high accuracy on KITTI, as demonstrated by 3DSSD~\citep{yang20203dssd}, PartA2~\citep{shi2019parta2}, and Voxel R-CNN~\citep{deng2020voxelrcnn}. However, weak LiDAR reflections for distant or occluded objects degrade detection performance~\citep{wallace2020full,li2020happens}.

\vspace{1mm}
\textbf{Multimodal Fusion Method.} Combining LiDAR and camera RGB input enhances detection performance, especially for sparse LiDAR objects~\citep{ma2019self}. Fusion approaches include projection-based methods, which align image features with point clouds in 3D space~\citep{huang2020epnet}, and model-based techniques that incorporate cross-attention~\citep{li2022deepfusion} or unified feature spaces~\citep{li2022uvtr}. Recent multimodal models, such as PPF-Det~\citep{xie2024ppfdet} and RoboFusion~\citep{song2024robofusion}, achieve top accuracy on KITTI benchmarks. However, these models often trade off accuracy for speed or vice versa.

\begin{figure*}[htbp]
    \centering
    \includegraphics[width=\linewidth]{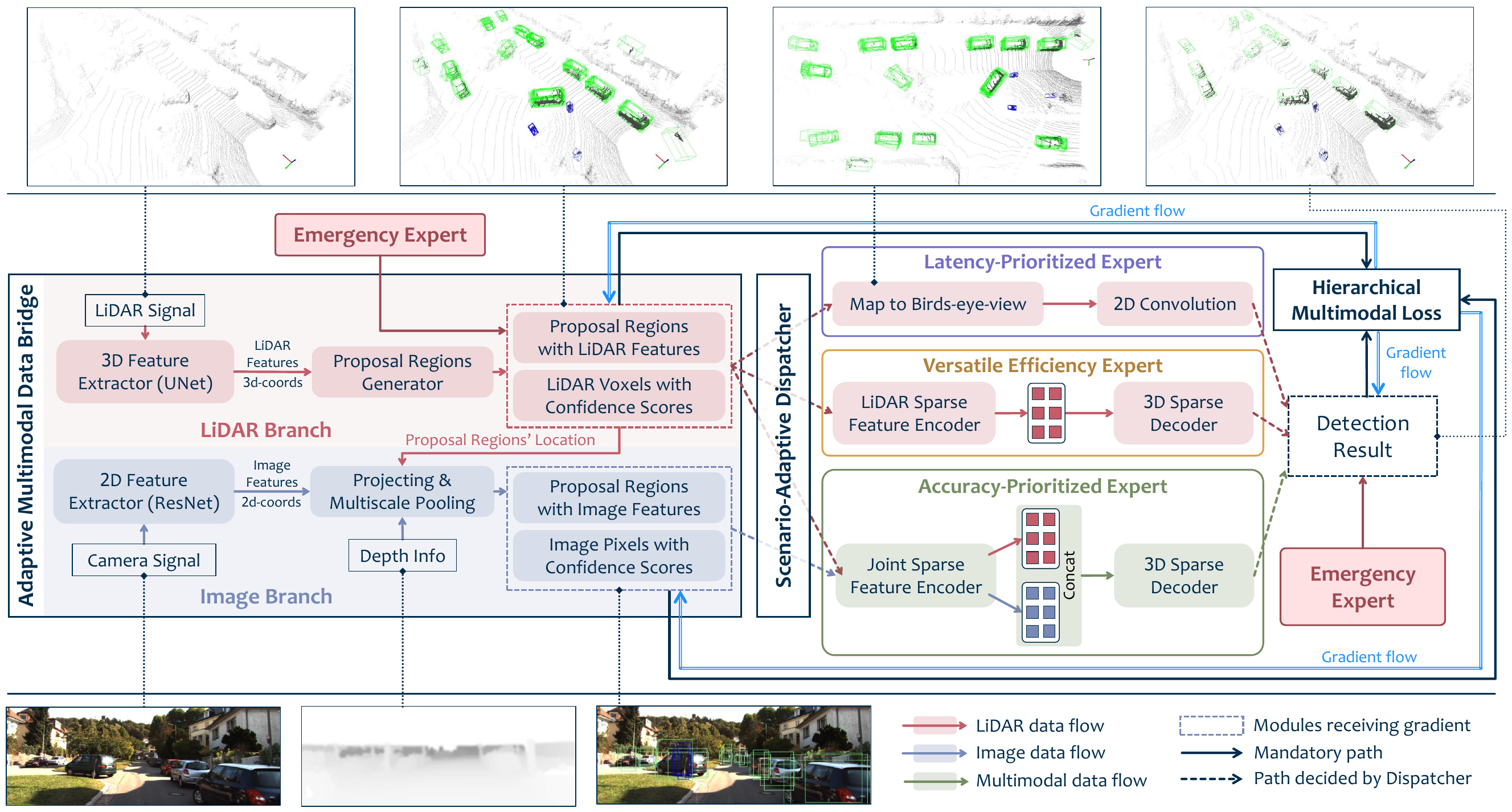}
    \vspace{-6mm}

    \caption{\textbf{System Structure of $\mathrm{EMC2}$.} The central row illustrates the architecture of $\mathrm{EMC2}$, while the top and bottom rows provide visual examples of core modules and relevant legends. The system integrates five main components: an \textit{Adaptive Multimodal Data Bridge (AMDB)} for preprocessing raw multimodal input and generating expert-specific features; a \textit{Scenario-Adaptive Dispatcher} that dispatches suitable experts based on scenario-specific features (see Fig.~\ref{fig:routing}); the \textit{Latency-Prioritized Expert} designed for simple scenarios requiring minimal inference delay; the \textit{Versatile Efficiency Expert} designed for uncertain or distant object detection scenarios; and the \textit{Accuracy-Prioritized Expert} responsible for handling complex, high-precision detection scenarios. Additionally, an \textit{Emergency Expert API} is designed to support rapid response to hazardous or unseen situations. The overall training process incorporates loss terms for both final detection results and intermediate outputs from the \textit{AMDB}, as described in Sec.~\ref{sec:train}.}
    \vspace{-4mm}
    \label{fig:architecture}
\end{figure*}

\vspace{1mm}
\textbf{Edge-based Compilation.} ML compilation platforms are essential for improving the inference efficiency of deep learning models across diverse hardware architectures~\citep{wang2018machine,leather2014automatic}. TensorRT~\citep{nvidia_tensorrt} is an NVIDIA inference optimizer that accelerates ML models on GPUs through layer fusion, precision calibration, and kernel auto-tuning. XLA~\citep{XLA2020} is a domain-specific compiler for TensorFlow enhancing efficiency by fusing operations and optimizing computation graphs. Open Neural Network Exchange (ONNX) Runtime~\cite{onnxgithub} provides an open standard for model interoperability, enabling efficient execution across CPUs, GPUs, and accelerators. TVM~\citep{chen2018tvm} is an open-source deep learning compiler that automates model tuning and tensor scheduling for efficient deployment across CPUs and GPUs.

\begin{figure}[t]
  \centering
  \includegraphics[width=\linewidth]{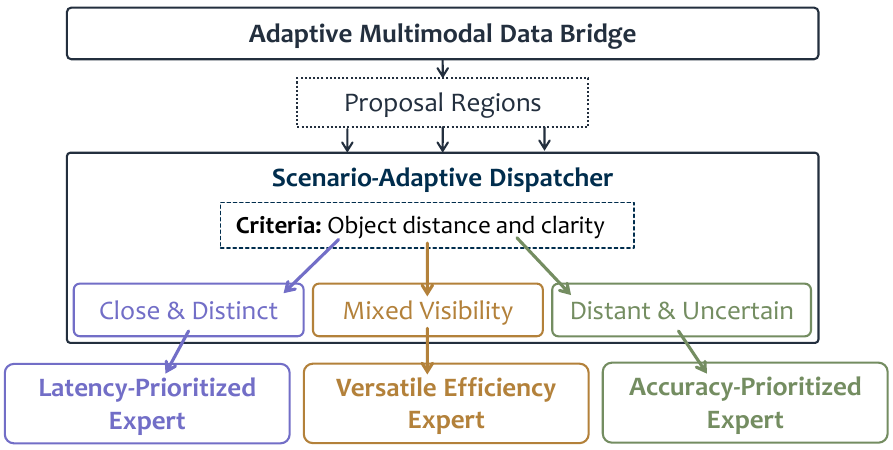}
  \vspace{-5mm}
   \caption{\textbf{Routing Strategy of the \textit{Scenario-Adaptive Dispatcher.}} The Routing criteria is detailed mathematically in Sec.~\ref{sec:sad}. }
   \label{fig:routing}
\end{figure}
\vspace{-1.5mm}
\section{EMC2 Computing Methodology}
\label{sec:method}
We introduce $\mathrm{EMC2}$, a multimodal MoE framework that minimizes latency and ensures reliable accuracy. Fig.~\ref{fig:architecture} shows the $\mathrm{EMC2}$ overview, $\mathrm{EMC2}$ consists of: 1) \textit{Adaptive Multimodal Data Bridge (AMDB)} to preprocess multimodal input (Sec.~\ref{sec:amdb}); 2) \textit{Scenario-Adaptive Dispatcher} to adaptively dispatch a suitable expert for each scenario (Sec.~\ref{sec:sad}); and 3) Three \textit{Scenario-Optimized Experts} to decode \textit{AMDB}-output features into final results (Sec.~\ref{sec:experts}). We also design an effective training strategy to address the long-tail effect in the training process (Sec.~\ref{sec:train}).In addition, we deploy $\mathrm{EMC2}$ on edge devices with self-designed 3D sparse convolution and Multiscale Pooling. 

\subsection{Adaptive Multimodal Data Bridge}
\label{sec:amdb}
Given the multimodal nature of $\mathrm{EMC2}$'s sensor inputs, a module providing suitable preprocessed data for different experts is required. Thus, we propose \textit{AMDB}, a preprocessing unit to efficiently process the required input data for each expert. The workflow of \textit{AMDB} begins with a UNet~\citep{ronneberger2015u} and a sparse CNN to extract LiDAR features, followed by fully connected layers that generate proposal regions along with their confidence scores. These results serve as inputs to \textit{Latency-Prioritized Expert} and \textit{Versatile Efficiency Expert}. The extraction of image features depends on the \textit{Scenario-Adaptive Dispatcher}. If further information is required for reliable detection, \textit{AMDB} will extract image features by ResNet. Based on depth information, it projects relevant image pixels into 3D space, applying Multiscale Pooling to reduce computational overhead, and fuses them with LiDAR voxel features to produce a multimodal output. These results are inputs for \textit{Accuracy-Prioritized Experts}. The Multiscale Pooling and multimodal fusion processes are demonstrated in Fig.~\ref{fig:pool_fuse}.

\vspace{-1mm}
\subsection{Scenario-Adaptive Dispatcher}
\label{sec:sad}
The \textit{Scenario-Adaptive Dispatcher (SAD)} dynamically dispatches each scenario to the most suitable expert based on traffic conditions and latency-accuracy requirements. Our expert selection algorithm considers real-world traffic safety needs: specifically, \uline{how far away objects are and how clearly they can be perceived.}

So, the switch among experts is determined by two key parameters: object distance $\mathcal{D}$ and clarity, which is inferred from the confidence $\mathcal{C}$ of proposal regions provided by the \textit{AMDB}. We classify traffic scenarios into three categories: \uline{1) Close and Distinct Cases.} All objects are clearly visible and no distant objects are present. This is characterized by all proposal regions $< \mathcal{D}$ and having confidence $\geq \mathcal{C}$. In such scenarios, as near-field objects provide abundant LiDAR information, even when compressed into bird’s eye view (BEV), a 2D CNN can effectively capture key patterns such as object surfaces and heights~\cite{alex2019pointpillars}, ensuring reliable detection accuracy with minimal latency (Sec.~\ref{sec:ablation}). These cases will be routed to the \textit{Latency-Prioritized Expert}. \uline{2) Mixed Visibility Cases.} Some objects are either distant but clearly visible or near but unclear. Mathematically, this corresponds to cases where i) some proposal regions $< \mathcal{D}$ with confidence $< \mathcal{C}$, or ii) the distance of some objects $\geq \mathcal{D}$ with confidence $\geq \mathcal{C}$. Here, objects have insufficient voxel data, requiring more complex 3D convolution to extract meaningful patterns, and they will be dispatched to the \textit{Versatile Efficiency Expert}. \uline{3) Distant and Uncertain Cases.} Some objects are both distant and unclear. This occurs when some proposal regions $\geq \mathcal{D}$ with confidence $< \mathcal{C}$. In these challenging scenarios, significant voxel information is missing due to long distances, steep reflection angles, or occlusions, necessitating the integration of image data to compensate for missing LiDAR details. These cases will be processed by the \textit{Accuracy-Prioritized Expert}.

\subsection{Scenario-Optimized Experts}
\label{sec:experts}


\textbf{Latency-Prioritized Expert (\textit{LPE}).} \textit{LPE} uses 2D CNNs for object detection based on BEV projections. It first projects proposal regions from the \textit{AMDB} into BEV space, then extracts 2D features for object localization and classification. It finally maps detection results back to 3D space. Compared to sparse 3D convolutions, its use of 2D operations reduces computation in outdoor ADS scenarios~\cite{alex2019pointpillars}, allowing \textit{LPE} to improve inference efficiency with minimal accuracy loss.


\textbf{Versatile Efficiency Expert (\textit{VEE}).} \textit{VEE} uses a 3D CNN to process proposal regions and confidence scores from the \textit{AMDB}. After sparse convolution, voxel features are decoded into per-voxel classifications and bounding boxes. Compared to \textit{LPE}, 3D convolutions better preserve spatial structure, making it more robust for scenarios with insufficient LiDAR signal.

\begin{figure}[t]
  \centering
  \includegraphics[width=\linewidth]{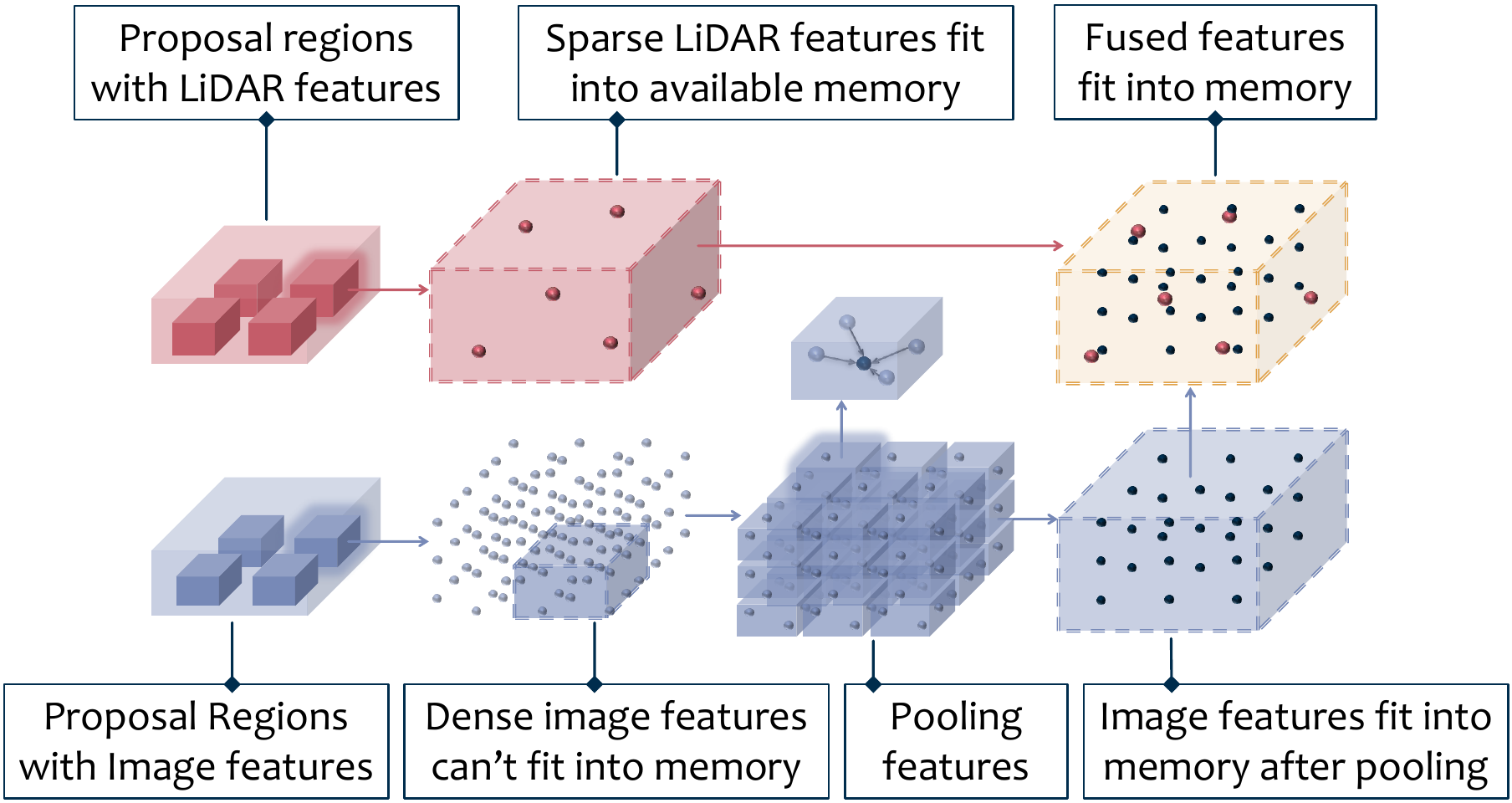}
   \vspace{-5mm}
   \caption{\textbf{Illustration of Multiscale Pooling and Multimodal Fusion.} 
   Proposal regions with image features are first obtained from the 3D proposal space generated by the LiDAR branch of \textit{AMDB}. After extracting the relevant image features from the corresponding image regions and projecting them into 3D space, pooling and fusion are then applied. Following the pooling step, image features of pixels that share the same 3D coordinate as a voxel are concatenated with the corresponding LiDAR voxel features, while those without matching voxel coordinates are concatenated with zero-filled features.}
   \vspace{-0.2in}
   \vspace{-1mm}
   \label{fig:pool_fuse}
\end{figure}

\textbf{Accuracy-Prioritized Expert (\textit{APE}).} \textit{APE} extends \textit{VEE} by incorporating the multimodal signal. It employs a 3D CNN and decoder with the same structure but different parameters to process multimodal proposal regions (from \textit{AMDB}) and fuses image data to recover missing LiDAR information. Image features compressed via Multiscale Pooling (Sec.\ref{sec:algOpt}) are projected into 3D space and fused with corresponding voxels of LiDAR features, ensuring more robust understanding for occluded and distant objects while keeping computation feasible. The Multiscale Pooling and multimodal fusion are detailed in Fig.~\ref{fig:pool_fuse}.

\begin{figure}[ht]
    \centering
    \includegraphics[width=0.73\linewidth]{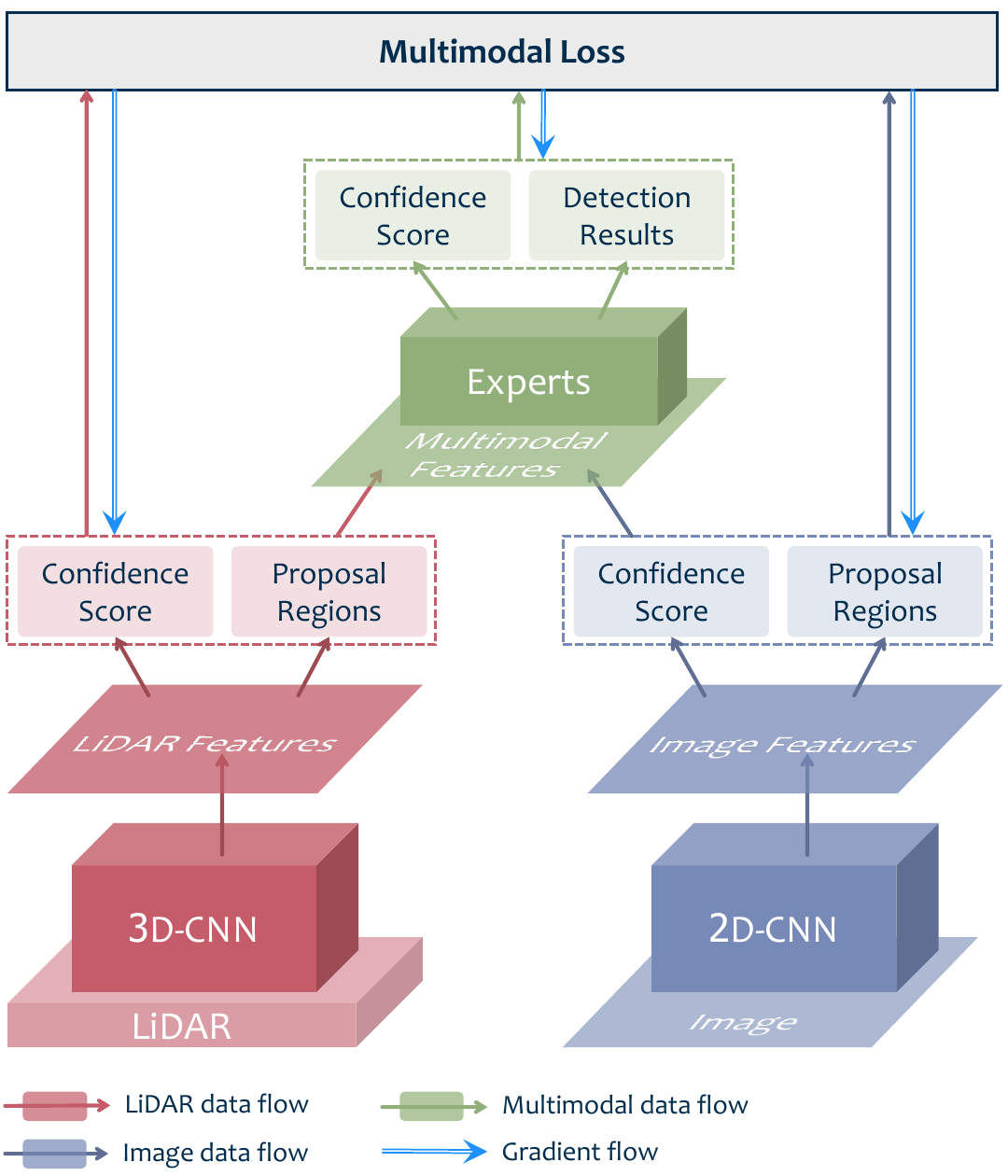}
    \vspace{-2mm}
    \caption{\textbf{Illustration of Hierarchical Training.} The three back-propagation paths originate from: 1) the final detection output, 2) the LiDAR branch output of \textit{AMDB}, and 3) the image branch output of \textit{AMDB}. Since expert training relies on initially unreliable \textit{AMDB} outputs, direct supervision accelerates convergence and enhances representation learning.}
    \label{fig:training}
    \vspace{-3mm}
\end{figure}

\subsection{Collaborative Training of Multimodal Experts}
\label{sec:train}

\textbf{Hierarchical Training Strategy: Multimodal Supervision with Triple Back-Propagation.} During training, the performance of experts is highly dependent on the proposal regions produced by the \textit{AMDB}, yet these proposals tend to be unreliable in the early training phase, leading to an unstable expert learning process and unstable \textit{SAD} decisions. To address this, we adopt a hierarchical supervision strategy with three distinct back-propagation pathways, as illustrated in Fig.~\ref{fig:training}: 1) supervision of the LiDAR branch outputs from the \textit{AMDB}; 2) supervision of the image branch outputs from the \textit{AMDB}; and 3) supervision of expert predictions and associated confidence scores. To further improve training stability, the LiDAR and image branches of the \textit{AMDB} are pre-trained separately before joint optimization. Each back-propagation route is supervised by the following loss function, which collectively forms Multimodal Loss:
\begin{align}
\mathcal{L}_\text{route}=\mathcal{L}_\text{cls}+\mathcal{L}_\text{reg}
    \label{eq:loss}
    \vspace{-0.1in}
\end{align}
where $\mathcal{L}_\text{cls}$ uses cross-entropy loss and $\mathcal{L}_\text{reg}$ uses smooth-$\ell1$ loss. The Hierarchical Training not only stabilizes expert learning but also enhances the model’s adaptability to various data granularities. In turn, even under compressed data representations, it enables the effective use of our Multiscale Pooling for memory efficiency, as detailed in Sec.~\ref{sec:algOpt}.

\textbf{Addressing Long-tail Effects in MoE Training.} The MoE framework often suffers from long-tail effects, where certain experts receive insufficient training data, leading to imbalances in sub-datasets, and therefore affecting performance~\cite{jacobs1991adaptive}. To address this, we propose a threefold training strategy: 1) \uline{Data subset division.} We partition the dataset into sub-datasets based on target and auxiliary samples. Each expert $\mathcal{E}_i$ is dispatched a sub-dataset $\mathcal{S}_i$ ($i=1,2,3$) containing target samples $\mathcal{T}_{\mathcal{S}_i}$, which are expected to be routed to the corresponding expert, and auxiliary samples $\mathcal{A}_{\mathcal{S}_i}$, which are not. We ensure that $\mathcal{T}_{\mathcal{S}_i} \cap \mathcal{T}_{\mathcal{S}_j} = \emptyset$ for any $i \neq j$. The inclusion of auxiliary samples, which are not intended for a given expert, helps prevent overfitting and enhances generalization. 2) \uline{Balanced sampling strategy.} We adjust the selection probability $\mathcal{P}_i$ for the sub-datasets which have fewer samples to ensure balanced training:

\vspace{-2mm}
   \begin{equation}
      \mathcal{P}_i \times \mathcal{N}_{\mathcal{S}_i} = \mathcal{P}_j \times \mathcal{N}_{\mathcal{S}_j},\;\, \sum_{i=1}^{3} \mathcal{P}_i = 1
   \end{equation}
where $\mathcal{N}_{\mathcal{S}_i}$ represents the number of samples in sub-datasets $\mathcal{S}_i$. This ensures that experts dispatched to less frequent scenarios receive sufficient training iterations. 3) \uline{Adaptive optimizer.} We introduce an optimizer that adapts the learning rate based on the proportion of target samples in a batch. Denoting $\alpha_0$ as the base learning rate, the adaptive learning rate applied to an expert with a given batch is computed as:
   \begin{equation}
      \alpha = \left( 1 + \frac{1}{\mathcal{N}} \sum_{i=1}^{\mathcal{N}} \mathbf{1}_{i \in \mathcal{T}} \right)\cdot\alpha_0
      \label{eq:lr}
   \end{equation}
where $\mathcal{N}$ is the batch size, $p$ is the proportion of target samples within current mini-batch, $\mathcal{T}$ represents target samples in the batch. By regulating the gradient update rate based on the batch composition, this adaptive optimizer encourages each expert to prioritize its designated target samples while preventing excessive updates from auxiliary samples.

\subsection{Algorithm Edge-adaptivity Optimization}
\label{sec:algOpt}
Edge devices face challenges such as incomplete function libraries and limited cache. We adopt the open-source ONNX Runtime for compiling $\mathrm{EMC2}$ on edge, which provides modules that help address these challenges and facilitate edge-oriented optimization. As 3D convolutions account for the majority of computation during inference, we customize \uline{3D sparse convolution} and \uline{Multiscale Pooling} to improve the algorithmic efficiency of $\mathrm{EMC2}$.

\vspace{1mm}

\textbf{3D Sparse Convolution.} Since the 3D sparse convolution library is incomplete on most edge platforms, we develop a customized 3D sparse convolution library with parallel threads execution under ONNX Runtime. Let $\mathcal{H}$ denote spatial resolution, $\mathcal{N}$ the number of non-empty voxels, and $\mathcal{C}{\text{v}}$ the per-voxel computation, including channels and kernel size. By computing nonzero voxels only, 3D sparse convolution reduces complexity from $O(\mathcal{H}^3\times \mathcal{C}_{\text{v}})$ for dense convolutions to $O(\mathcal{N}\times \mathcal{C}_{\text{v}})$, where $\mathcal{N} \ll \mathcal{H}^3$, cutting redundant operations by 65–80\%~\cite{yan2018second}. 

\vspace{1mm}
\textbf{Multiscale Pooling.} To alleviate cache limitations, Multiscale 3D Sparse Pooling adaptively pools image features based on available memory before fusing them with LiDAR features (Fig.\ref{fig:pool_fuse}). The pooling size is user-configurable, allowing flexibility for maximizing memory utilization. When combined with Hierarchical Training (Sec.\ref{sec:train}), Multiscale Pooling effectively reduces resource consumption while maintaining model reliability (Sec.~\ref{sec:ablation}).

\subsection{EMC2 Computing System Optimization}
\label{sec:sysOpt}
Building upon the algorithm-level optimizations introduced in Sec.~\ref{sec:algOpt}, we further enhance the ONNX execution efficiency through system-level improvements, specifically targeting \uline{memory management} and \uline{computation graph optimization}.

\begin{figure*}[htbp]
    \centering
    \includegraphics[width=0.9\linewidth]{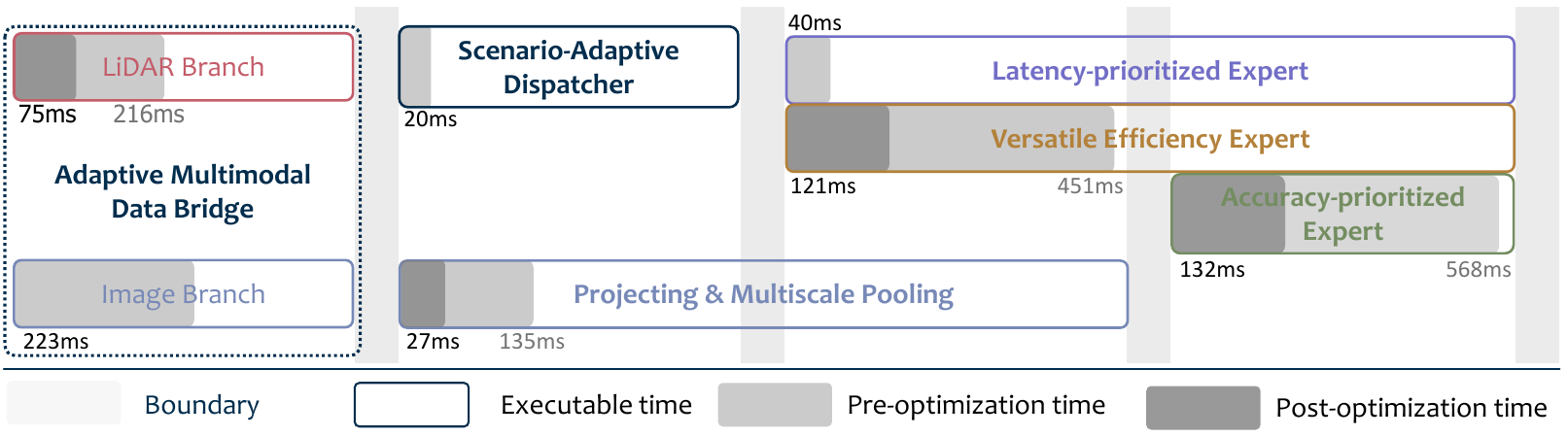}
    \vspace{-2mm}
    \caption{\textbf{Illustration of Thread Management.} Each solid box represents a module occupying threads. During inference, four boundaries are defined at which preceding threads are terminated, and their thread-specific address space is released upon crossing each boundary.}
    \vspace{-0.1in}
    \label{fig:thread}
\end{figure*}

\vspace{1mm}
\textbf{Memory Optimization.} To further improve memory efficiency during inference, we implement three system-level techniques. \uline{1) Overlapping communication and computation~\cite{hromkovivc2013communication}.} Matrix data is partitioned into segments processed and transferred in parallel, overlapping computation and communication. This reduces global memory access, improves cache usage, and mitigates memory bottlenecks. \uline{2) Thread scheduling.} We implement a staged thread management mechanism, where active threads are periodically terminated and re-initialized (Fig.~\ref{fig:thread}). This approach limits unnecessary memory retention between stages, reducing system memory footprint. \uline{3) Prefix-sum for sparse convolution.} Parallel prefix-sum reduces index search time from $\mathcal{O}(\mathcal{N})$ to $\mathcal{O}(\log \mathcal{N})$ using GPU parallelism, where $\mathcal{N}$ is the number of nonzero elements.

\textbf{Computational Graph Optimization.} ONNX partitions the computation graph into segments and assigns each to the most suitable hardware. Our optimization includes three stages: \uline{1) Model pruning.} Redundant parameters and operations are removed to simplify the graph; \uline{2) Model quantization.} Low-precision arithmetic replaces floating-point computation to improve hardware efficiency; \uline{3) Computational graph fusion.} Consecutive operators are merged to reduce memory access and improve execution, benefiting memory-bound CPUs and reducing kernel launch overhead on GPUs.

\vspace{-1mm}
\section{Experiment and Result}
\label{sec:experiments}



\vspace{-1mm}

\subsection{Experiment Settings and Datasets}
\label{sec:setting}
\begin{table*}[htbp]
\footnotesize
\setlength{\tabcolsep}{9.5pt}
\centering
\begin{tabular}{cccccccccccc}
\Xhline{1pt}
\multirow{3}{*}{\textbf{Method}} & \multicolumn{3}{c}{\rule{0pt}{1.1em}\textbf{Pedestrian 3D AP (R40) $\uparrow$}} & \multicolumn{3}{c}{\textbf{Car 3D AP (R40) $\uparrow$}} & \multicolumn{3}{c}{\textbf{Cyclist 3D AP (R40) $\uparrow$}} & \multicolumn{2}{c}{\textbf{Latency (ms)} $\downarrow$} \\
\cmidrule(lr){2-4} \cmidrule(lr){5-7} \cmidrule(lr){8-10} \cmidrule(lr){11-12}
 & \textbf{Easy} & \textbf{Mod.} & \textbf{Hard} & \textbf{Easy} & \textbf{Mod.} & \textbf{Hard} & \textbf{Easy} & \textbf{Mod.} & \textbf{Hard} & \textbf{A4000} & \textbf{Jetson} \\
\midrule
\vspace{-0.4cm}\\
\multicolumn{12}{c}{\textbf{LiDAR-only Models}}\\
\hline
\hline
M3DETR~\citep{guan2022m3detr} & 69.69 & 66.04 & 61.61 & 93.96  & 86.28 & 84.32 & 87.88 & 72.47 & 70.63 & 338 & N/A \\ 
PV-RCNN~\citep{shi2020pv} & N/A & N/A & N/A & 91.18 & 87.65 & 83.14 &  N/A & N/A & N/A & 45 & 1787 \\
Voxel-RCNN~\citep{deng2020voxelrcnn} & N/A & N/A & N/A &92.23 & 85.04 & 82.50 &  N/A & N/A & N/A & 70 & N/A \\
GLENet-VR~\citep{liao2022gle} & N/A & N/A & N/A & 93.51 & 86.22 & 83.72 &  N/A & N/A & N/A & 165 & N/A \\
PartA2-Anchor~\citep{shi2020points} & 66.87 & 59.68 & 54.60 & 92.15 & 82.92 & 82.10 & 90.34  & 70.05 & 66.89 & 95 & N/A \\
PartA2-Free~\citep{shi2020points} & 72.31 & 66.36 & 60.06 & 91.66 & 80.28 & 78.08 & 91.88 & 75.33 &  70.67 & 124 & 1064 \\
PointPillar~\citep{alex2019pointpillars} & 57.29  & 51.41 & 46.87  &87.75 & 78.40 & 75.19 & 81.57 &  62.93 & 58.97  & \textbf{41} & 972 \\
Second~\citep{yan2018second} & 55.94  &51.14 & 46.16 &90.55 & 81.60 & 78.60 &82.96 & 66.73 & 62.78 &  45 & 1322 \\
Second-IoU~\citep{yan2018second} & 61.10& 54.66 & 49.50 & 91.53 & 82.36 & 79.62 &90.73 & 71.23  & 66.26 &  58 & N/A \\
3ONet~\citep{hoang20233onet} & 72.55 & 65.21 & 60.22 & \textbf{94.24} & 87.32 & 84.17 & 92.47 & 75.11 & 71.18 & 100 & N/A \\
\midrule
\vspace{-0.4cm}\\
\multicolumn{12}{c}{\textbf{Multimodal Models}}\\
\hline
\hline
Voxel-RCNN~\citep{deng2020voxelrcnn} & N/A & N/A & N/A &92.08 &  85.90 & 83.36 & N/A & N/A & N/A & 331 & 965 \\
LoGoNet~\citep{li2023logonet} & 70.02 & 63.72 & 59.46 & 92.04 & 85.04 & 84.31  & 91.74 & 75.35 & 72.42 & 100 & N/A \\
VFF~\citep{li2022vff} & 73.26 & 65.11 & 60.03 & 92.24 & 85.51 & 82.92 & 91.74 & 75.35 & 69.84 & 192 & N/A \\
CAT-DET~\citep{zhang2022cat} & 74.08 & 66.35 & 58.92 & 90.12 & 81.46 & 79.15 & 87.64 & 72.82 & 68.20 & 154 & N/A \\
TED~\citep{wu2023transformation} & 74.73 & 69.07 & 63.63 & 92.25 & 88.94 & 86.73 & \textbf{95.20} & 76.17 & 71.59 & 162 & N/A \\
\rowcolor{blue!10} $\mathrm{EMC2}$ (Ours) &\textbf{74.92}  &\textbf{69.92}  & \textbf{66.81}  & 94.00&\textbf{89.35} &\textbf{88.15}  &95.06  &  \textbf{79.87} &  \textbf{77.62} & 160 & \textbf{372.5} \\
\specialrule{1pt}{0pt}{0pt}
\end{tabular}
\vspace{-2mm}
\caption{We evaluate different 3D object detection methods on the KITTI dataset. Some cells are denoted as \textit{N/A} when the corresponding open-source implementation lacks support for the specific object class or is not compatible with Jetson devices.}
\label{tab:comparison}
\vspace{-4mm}
\end{table*}

\begin{table}[t]
\centering
\footnotesize
\captionsetup{font=small, skip=1pt}
{\begin{tabular}{lccccc}
\toprule
\textbf{Method} & \textbf{Size} & \textbf{mAP}$\uparrow$ & \textbf{NDS}$\uparrow$ & \textbf{Latency (ms)}$\downarrow$  \\
\midrule
FocalFormer~\cite{chen2023focalformer3d} & 189M & {0.6640} & 0.7090 & N/A \\
BEVFusion~\cite{liu2023bevfusion}& {156M} & {0.6852} & {0.7138} & N/A \\
\rowcolor{blue!10} 
$\mathrm{EMC2}$ (ours) & 87M & \textbf{0.7241} & \textbf{0.7316} & \textbf{229.3} \\
\bottomrule
\end{tabular}}
\vspace{2mm}
\caption{Comparison of different methods on the nuScenes dataset.}
\label{tab:nuscenes_exp}
\vspace{-1mm}
\end{table}

\textbf{Experiment Platform.} Experiments are conducted on the Jetson AGX Orin, an edge platform with limited computational resources, an NVIDIA Ampere GPU, a 4MB shared L2 cache, and hierarchical memory. For comparison, $\mathrm{EMC2}$ is also evaluated on an NVIDIA A4000 GPU.

\vspace{1mm}
\textbf{Dataset and Sample Difficulties.} We evaluate 3D object detection on two public benchmarks: KITTI~\cite{andreas2012kitti} and nuScenes~\cite{caesar2020nuscenes}. KITTI provides 3,712 training and 3,769 validation samples, with object difficulty levels categorized as 30\% easy, 40\% moderate, and 30\% hard. nuScenes contains 850 training and 150 test scenes, covering 10 object categories with 3D bounding box annotations. Before splitting the data, we analyze the confidence distribution of proposal regions generated by the pre-trained \textit{AMDB}. A two-sample Kolmogorov–Smirnov test reveals a statistically significant difference between objects within and beyond a distance threshold $D$, which is empirically set to $23.5$ meters for KITTI and $35$ meters for nuScenes. The training subsets for each expert on the KITTI and nuScenes datasets are defined as follows: $\mathcal{T}_{\mathcal{S}_1}$ includes scenes where all objects are within $D$ meters and labeled as “easy” or “moderate”; $\mathcal{T}_{\mathcal{S}_2}$ includes scenes with at least one object beyond $D$ meters labeled as “easy” or “moderate,” or any object within $D$ meters labeled as “hard”; and $\mathcal{T}_{\mathcal{S}_3}$ includes scenes where at least one object is both beyond $D$ meters and labeled as “hard,” requiring multimodal processing. For each expert, the auxiliary subset $\mathcal{A}_{\mathcal{S}_i}$ is uniformly sampled from the other experts’ target subsets, with its size equal to that of the corresponding target subset.

\vspace{1mm}
\textbf{Evaluation Metrics.} KITTI uses R40 3D Average Precision (AP) with IoU thresholds of 0.7 for cars and 0.5 for pedestrians and cyclists. nuScenes adopts the NDS metric, combining mAP—computed via center distance matching (0.5\,m to 4.0\,m thresholds)—and five error terms: mean Average Translation (mATE), Scale (mASE), Orientation (mAOE), Velocity (mAVE), and Attribute (mAAE) Errors. Inference latency on Jetson AGX Orin is also reported to assess real-time performance.

\label{sec:impleDetails}
\textbf{Hierarchical Training Details.} Allowing all experts to propagate gradients to \textit{AMDB} causes instability due to overlapping training data, leading to repeated updates. To address this, \textit{AMDB} is first pre-trained for 20 epochs, followed by joint training where gradients are propagated only from \textit{APE}, which receives full multimodal input.

\subsection{Comparison with Existing SOTA Methods}
\label{sec:comparison}

As shown in Tab.~\ref{tab:comparison} and Tab.~\ref{tab:nuscenes_exp}, $\mathrm{EMC2}$ achieves SOTA accuracy and latency on KITTI and nuScenes. On the Jetson AGX Orin platform, it also outperforms existing baseline models across all difficulty levels.

\vspace{1mm}
\textbf{Results on KITTI.} Pedestrians and cyclists face higher traffic risks~\cite{lubbe2022safe}, which remain challenging for most SOTA methods. $\mathrm{EMC2}$ improves hard-set detection by $5\%$ and $7\%$ for pedestrians and cyclists, respectively, outperforming TED and LoGoNet (Tab.~\ref{tab:comparison}). For cars, it achieves $88.15\%$ on the hard set, exceeding prior multimodal methods by up to $9\%$. It runs at $372.5$ ms on Jetson, $2.5\times$–$2.5\times$ faster than methods requiring $965$–$1,787$ ms.

\vspace{1mm}
\textbf{Results on nuScenes.} $\mathrm{EMC2}$ improves mAP and NDS by $3.9\%$ and $1.8\%$, respectively, outperforming recent methods FocalFormer~\cite{chen2023focalformer3d} and BEVFusion~\cite{liu2023bevfusion} (Tab.~\ref{tab:nuscenes_exp}). It runs at $229.3$ ms on Jetson AGX Orin, enabling real-time use on edge devices.

\begin{figure}[t]
\captionsetup{font=small, skip=4pt}
  \centering
  \includegraphics[width=1\linewidth]{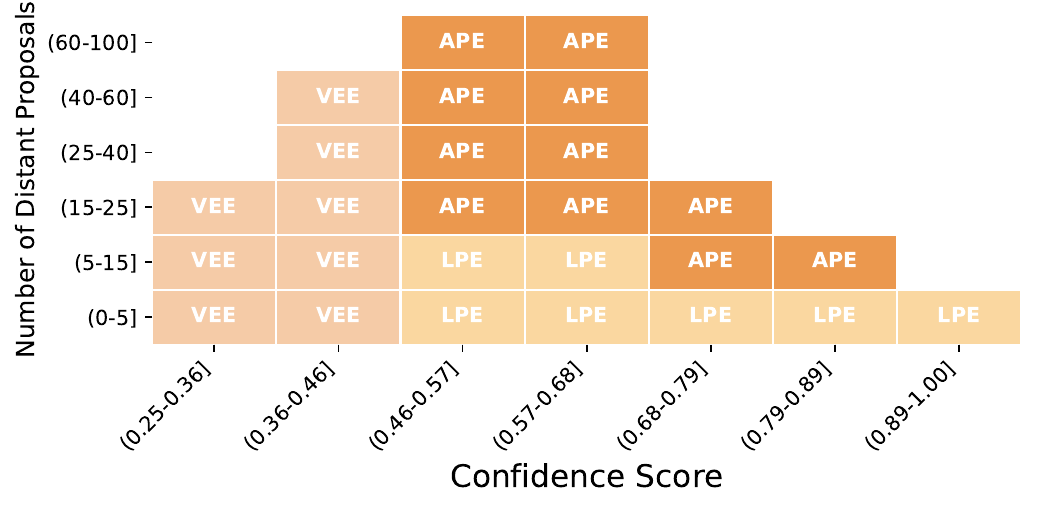}
   \vspace{-5mm}
   \caption{\textbf{Distribution of expert activation.} During $\mathrm{EMC2}$ inference on the nuScenes validation set, scenes are grouped into a 2D grid based on two metrics: the y-axis indicates the number of \textit{AMDB}-generated proposals beyond the distance threshold $D$ (i.e., distant objects), and the x-axis represents the average confidence score of these proposals (higher values indicate clearer scenes). Each cell is labeled with the name of the expert that performs best for scenes falling into that bin.}
   \label{fig:best_expert_grid}
   \vspace{-3mm}
\end{figure}

\subsection{Analysis and Discussion}
\textbf{Benefits of Mixture-of-Experts.} The balance between accuracy and efficiency observed in Sec.~\ref{sec:comparison} is achieved through dynamic expert selection based on scenario characteristics, combined with algorithmic and system-level optimizations designed for edge devices, such as Jetson.

\vspace{1mm}
\textbf{Empirical Dispatcher.} The $\mathrm{EMC2}$ dispatcher uses fixed distance and confidence thresholds to dispatch scenes to experts. The protocol is adapted per dataset format. This strategy is validated via expert activations on the nuScenes validation set (Fig.~\ref{fig:best_expert_grid}): \textit{LPE} performs best in close-range, high-confidence scenes; \textit{APE} in distant, low-confidence cases; and \textit{VEE} in all others, supporting the effectiveness of empirical routing.

\vspace{1mm} 
\textbf{Optimized Parameter Size.} $\mathrm{EMC2}$ has 226M parameters, but its MoE framework activates only a subset during inference, averaging 87M on nuScenes and 146M on KITTI, which is fewer than those used by the full model and other SOTA baselines. System-level optimizations instead prioritize computational throughput and memory efficiency over parameter reduction.

\vspace{1mm}
\textbf{Hardware Flexibility.} The $\mathrm{EMC2}$ is implemented in the ONNX format, an open format supported by various compilers and enables model compilation across diverse hardware platforms. As a result, $\mathrm{EMC2}$ is not limited to Jetson and can be deployed on other edge platforms.

\begin{figure}[t]
  \centering
  \includegraphics[width=0.96\linewidth]{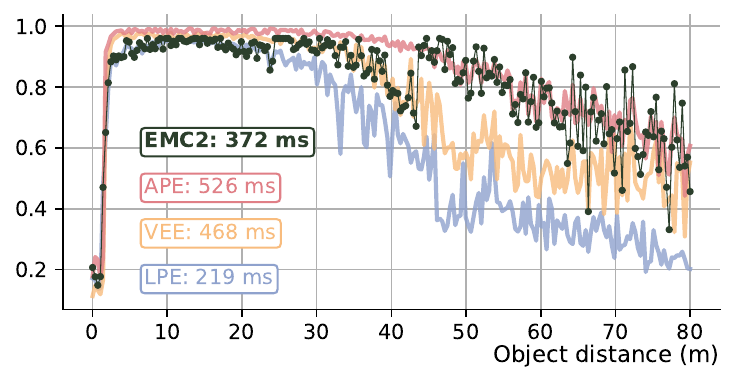}
  \vspace{-0.1cm}
  \caption{\textbf{Median Detection Confidence vs. Object Distances.} Four configurations are compared on the KITTI validation set: LPE (blue), VEE (peach), APE (rose), and $\mathrm{EMC2}$ (green). More complex experts yield higher confidence but slower inference. By switching experts based on distance and clarity, $\mathrm{EMC2}$ balances accuracy and efficiency.}
  \vspace{-2mm}
  \label{fig:score-distance}
\end{figure}

\begin{table}[t]
\footnotesize
\setlength{\tabcolsep}{7pt}
\begin{tabular}{ccccc}
\Xhline{1.2pt}
\small\multirow{3}{*}{\textbf{Method}} & \multicolumn{3}{c}{\rule{0pt}{1em}\textbf{AP} $\uparrow$} & \small\multirow{3}{*}{\textbf{Latency (ms)} $\downarrow$} \\
\cmidrule(lr){2-4}
 & \textbf{Ped.} & \textbf{Car} & \textbf{Cyclist} & \\
\midrule
\vspace{-0.35cm}\\
\multicolumn{5}{c}{\textbf{Ablation on \textit{LPE}}}\\
\rowcolor{blue!10}\textit{LPE} & 66.32 & 89.04 & 88.54 & \textbf{219} \\
\textit{VEE} & \textbf{74.49} & 92.23 & 91.69 & 468 \\
\textit{APE} &74.34 & \textbf{93.21} & \textbf{92.41} & 526 \\
\midrule
\vspace{-0.35cm}\\
\multicolumn{5}{c}{\textbf{Ablation on \textit{VEE}}}\\
\rowcolor{blue!10} W/ \textit{VEE} & \textbf{70.55} & \textbf{90.50} & \textbf{84.07} & \textbf{372.5} \\
W/O \textit{VEE} & 68.26 & 86.40 & 79.17 & 512.7 \\
\midrule
\vspace{-0.35cm}\\
\multicolumn{5}{c}{\textbf{Ablation on \textit{APE}}}\\
\textit{VEE} & 67.24 & 87.69 & 81.85 & \textbf{468} \\
\rowcolor{blue!10}\textit{APE} & \textbf{70.55} & \textbf{90.50} & \textbf{84.18} & 526 \\
\midrule
\vspace{-0.35cm}\\
\multicolumn{5}{c}{\textbf{Ablation on Hierarchical Training and Multiscale Pooling}}\\
\rowcolor{blue!10}HT \cmark, MP \cmark  & \textbf{70.55} & \textbf{90.50} & 84.07   & \textbf{526} \\
HT \cmark, MP \xmark  & \textbf{70.55} & \textbf{90.50} & \textbf{84.18}  & 1320 \\
HT \xmark, MP \cmark  & 63.11 & 88.68 & 75.98  & 529 \\
HT \xmark, MP \xmark  & 66.34 & 89.10 & 80.83  & 1367 \\
\Xhline{1.2pt}
\end{tabular}
\caption{\textbf{Results Across Three Ablation Configurations.} The main body of this table is divided into four sections corresponding to each part of the ablation study. In the third section, HT and MP are abbreviated for Hierarchical Training and Multiscale Pooling, respectively.}
\label{tab:ablation}
\vspace{-4mm}
\end{table}

\subsection{Ablation Study}
\label{sec:ablation}

\textbf{Ablation on the Roles of Three Experts.} We evaluate each expert on its target scenarios using the KITTI validation set. As shown in Tab.~\ref{tab:ablation}, \textit{LPE}, \textit{VEE}, and \textit{APE} achieve comparable accuracy in \textit{LPE} scenes, but \textit{LPE} has the lowest inference latency on Jetson. In \textit{VEE} scenarios, incorporating \textit{VEE} improves both accuracy and latency over the baseline. For \textit{APE}, which incorporates image features, accuracy improves further with only a minor latency increase, supported by the score distribution in Fig.~\ref{fig:score-distance}.

\vspace{1mm}
\textbf{Ablation on Hierarchical Training (HT) and Multiscale Pooling (MP).} We evaluate four combinations of HT and MP on the full KITTI validation set: both applied, HT only, MP only, and neither. Tab.~\ref{tab:ablation} shows that combining HT and MP reduces inference time on Jetson from $1320$ ms to $526$ ms without accuracy loss. For the car class, MP alone reduces accuracy from $80.83\%$ to $75.98\%$, while HT alone improves it from $66.34\%$ to $70.55\%$. HT facilitates multigranular feature learning, while MP enhances memory efficiency via compression.


\section{Conclusion}
\label{sec:conclusion}
\vspace{-1mm}

We present $\mathrm{EMC2}$, a multimodal 3D object detection solution for ADS based on a customized MoE architecture. This design enables adaptive expert selection across diverse traffic scenarios, balancing detection accuracy and efficiency. Experiments on KITTI and nuScenes show strong performance, particularly under challenging conditions. On the Jetson platform, $\mathrm{EMC2}$ achieves a $159.06\%$ speedup in inference, supporting real-time edge deployment. Future work will explore fully adaptive expert selection to enhance system responsiveness and resource efficiency.

\clearpage
{
    \small
    \normalem
    
}

\end{document}